\DocumentMetadata{
	lang		= eng-US, 	
	pdfstandard	= A-3u,		
	pdfversion	= 1.7, 		
}

\documentclass[colorlinks,upint,subscriptcorrection,varvw,hyphenate,balance,greek,russian,vietnamese,german, nohead, nofoot]{asmeconf} 

\usepackage{ulem}
\usepackage{siunitx}
\usepackage{float}
\usepackage{arydshln}
\usepackage{booktabs}
\usepackage{soul}
\usepackage{xcolor} 
\usepackage{amsmath}
\usepackage{algorithm}
\usepackage{algpseudocode}
\floatname{algorithm}{Algorithm}

\hypersetup{%
	pdfauthor={Kittipong Thiamchaiboonthawee},									
	pdftitle={FLARE},                 
	pdfkeywords={Additive Manufacturing, Directed Energy Deposition, Surrogate Modeling, Neural Implicit Fields, Field Prediction, Data-driven Learning},
	pdfsubject = {Data-efficient Surrogate Modeling},			

}

\allowdisplaybreaks

\begin{document}

\title{FLARE: A Data-Efficient Surrogate for Predicting Displacement Fields in Directed Energy Deposition}

\SetAuthors{%
    Kittipong Thiamchaiboonthawee\affil{1}\CorrespondingAuthor{kt455@mit.edu}, 
    Ghadi Nehme\affil{1},
    Ram Mohan Telikicherla\affil{2},
    Jiawei Tian\affil{2},
    Balaji Jayaraman\affil{2},\\
    Vikas Chandan\affil{2},
    Dhanushkodi Mariappan\affil{2},
	Faez Ahmed\affil{1}
	}

\SetAffiliation{1}{Massachusetts Institute of Technology, Cambridge, MA, USA}
\SetAffiliation{2}{GE Vernova Advanced Research Center, Niskayuna, NY, USA}

\maketitle
\pagestyle{plain}
\versionfootnote{This is a preprint. The contents are currently under review.}



\begin{abstract}
Directed energy deposition (DED) produces complex thermo-mechanical responses that can lead to distortion and reduced dimensional accuracy of a manufactured part. Thermo-mechanical finite element simulations are widely used to estimate these effects, but their computational cost and the complexity of accurately capturing DED physics limit their use in design iteration and process optimization. This paper introduces FLARE (Field Prediction via Linear Affine Reconstruction in wEight-space), a data-efficient surrogate modeling framework for predicting post-cooling displacement fields in DED from geometric and process parameters. We develop a predefined-geometry DED simulation workflow using an open-source finite element framework and generate a dataset of simulations with varying geometry, laser power, and deposition velocity. Each simulation provides full-field displacement, stress, strain, and temperature data throughout the manufacturing process. FLARE encodes each simulation as an implicit neural field and regularizes the corresponding neural-network weights so that they follow the affine structure of the input parameter space. This enables prediction of unseen parameter combinations by reconstructing network weights through affine mixing of training examples. On this DED benchmark, the method shows improved accuracy compared to baseline methods in both in-distribution and extrapolation settings. Although the present study focuses on DED displacement prediction, the proposed affine weight-space reconstruction framework offers a promising approach for data-efficient surrogate modeling of physical fields.
\end{abstract}







\section{Introduction}
Directed energy deposition (DED) is a metal additive manufacturing (AM) process in which focused thermal energy fuses material as it is deposited \cite{f422022terminology}. Despite its flexibility for fabrication and repair, DED induces severe thermal load that can produce plastic deformation, residual stresses, and part-scale distortion. These effects often limit dimensional accuracy and downstream part performance, motivating predictive tools that can estimate distortion before fabrication. However, DED simulations remain computationally expensive, creating a need for fast surrogate models that can support distortion prediction in design and process-planning workflows \cite{heigel2015thermo, stender2018thermal}.

Physics-based thermo-mechanical finite element simulations remain a primary approach for estimating temperature evolution and the resulting mechanical response in metal additive manufacturing \cite{michaleris2014modeling, heigel2015thermo, stender2018thermal}. Accurate estimation of displacement is important because build-induced distortion can determine whether a part satisfies dimensional requirements and can guide compensation, redesign, and process-parameter selection prior to fabrication. For example, an engineer may wish to vary geometric and/or process parameters of a part, but cannot readily assess how those changes will affect the resulting distortion without repeatedly running expensive simulations. Fast predictive capability is therefore especially valuable in iterative workflows that require repeated evaluation of candidate geometries or process settings, such as design exploration, compensation, and feasible parameter search. However, DED simulations must resolve moving heat sources, evolving domains (material activation/deposition), and temperature-dependent constitutive behavior, making them computationally expensive and difficult to embed in tasks such as optimal and feasible parameter search \cite{stender2018thermal, yushu2022directed}.

This need is further complicated by the data limitation of typical physics-based manufacturing problems. Simulations are expensive to generate, so datasets are often small, while the outputs of interest are spatial fields with nonlinear underlying physical phenomena. 
Publicly available full-field metal AM datasets are limited, especially for DED and full-field results (e.g. displacement, strain, stress, temperature fields). Recent publicly available datasets include a repository with thermal and displacement fields for Wire-arc Additive Manufacturing (WAAM) \cite{asociacion_de_investigacion_metalurgica_2025_17608626} and a dataset of simulated thermal fields for laser powder-fed DED \cite{chechik2025improved, chechik_2025_17454378}. We provide 164 simulations with full-field outputs spanning a seven-dimensional design space defined by five geometry variables and two process parameters. Each simulation includes the full displacement field on a mesh of approximately 32k nodes over roughly 500 time steps.

We introduce FLARE (\textbf{F}ield Prediction via \textbf{L}inear \textbf{A}ffine \textbf{R}econstruction in w\textbf{E}ight-space), a data-driven surrogate framework for controllable generation of physical fields guided by tabular parameters. FLARE represents each training sample’s displacement field as a neural implicit field. Rather than learning a single monolithic network mapping parameters to fields, FLARE learns a set of per-sample implicit fields whose weights are regularized to follow the same affine structure as the parameter space, enabling weight-space reconstruction for both interpolation and extrapolation queries. At inference, a target parameter vector is expressed as an affine combination of training parameters and the corresponding field is generated by the same combination in weight space, yielding a lightweight, controllable field generator.

We evaluate the proposed method on the generated thermo-mechanical DED dataset and demonstrate improved accuracy compared to parameter-conditioned neural field models and neural operator baselines. The method also enables fast inference suitable for design and process-planning workflows. In addition, we train a feasibility discriminator that identifies parameter combinations leading to nonphysical conditions, enabling integration with downstream optimization or generative design pipelines.

In summary, our key contributions are as follows:
\begin{itemize}
    \item \textbf{Affine Weight-Space Regularization:} We propose an affine weight-space regularization strategy that aligns the weight-space of the neural implicit field representation with input parameter-space, improving controllable field generation.
    \item \textbf{Data-Efficient Field Prediction:} We evaluate FLARE on thermo-mechanical DED simulation data and compare against parameter-conditioned baselines (conditional neural fields and neural operator \cite{perez2018film,lu2021learning}), demonstrating improved accuracy and robustness to diminishing data.
    \item \textbf{Open-Source Predefined-Geometry DED Simulation:} We provide an input template for simulating the DED process with prescribed geometries using the open-source MOOSE Application Library for Advanced Manufacturing UTilitiEs (MALAMUTE), which is built on the Multiphysics Object-Oriented Simulation Environment (MOOSE) framework \cite{malamute_github, permann2020MOOSE, lindsay20222, giudicelli20243, harbour20254}. Our approach is adapted from an existing geometry-free method \cite{yushu2022directed}. 
    MOOSE enables scalable and parallel workflows for large-scale dataset generation while avoiding licensing constraints and supporting deployment across HPC environments.
\end{itemize}

\section{Related Work}



This work considers surrogate modeling of process-induced displacement fields in DED in the post-cooling phase under varying geometry and process conditions. DED is an important additive manufacturing process for fabrication and repair, and fast, accurate prediction can accelerate the design cycle while ensuring part quality and performance \cite{denlinger2015effect, denlinger2016effect, blakey2021metal, svetlizky2022laser, chen2022review}. Because DED simulations are expensive and available datasets are limited, DED presents a challenging setting for data-efficient field prediction.

\subsection*{Thermo-mechanical modeling of DED}

Thermo-mechanical simulation of metal additive manufacturing typically couples transient heat conduction with quasi-static mechanical equilibrium, while material deposition is represented through element activation or other evolving-domain strategies \cite{michaleris2014modeling, stender2018thermal, yushu2022directed}. These models capture the interaction between a moving thermal input, developing geometry, and heat transfer, enabling prediction of residual stresses, distortions, and, depending on the length-scale of the simulation, the micro-structural evolution during fabrication.

Continuum-scale formulations can demonstrate how deposition-induced thermal histories drive residual stress formation and structural distortion. A body of literature highlights the sensitivity of predictions to heat source representations, boundary conditions, and discretization strategies \cite{heigel2015thermo, irwin2016line, yang2016finite}. Subsequent studies have validated such thermo-mechanical models against experimental measurements across multiple additive manufacturing processes, demonstrating their predictive capability at the part scale \cite{heigel2015thermo, denlinger2017thermomechanical, yang2016finite}.

Modern implementations emphasize scalable multiphysics frameworks and efficient domain construction strategies to support large-scale simulations \cite{stender2018thermal, permann2020MOOSE, lindsay20222, giudicelli20243, harbour20254}. Techniques such as geometry-free domain reconstruction and adaptive subdomain modeling have shown that choices in domain representation, adaptivity, and coupling can affect memory usage and computational throughput \cite{yushu2022directed}. Even with efficient multiphysics frameworks and workflow engineering, the cost of generating simulations for data-driven learning across geometry and process settings can be prohibitive \cite{permann2020MOOSE, lindsay20222, giudicelli20243, harbour20254, yushu2022directed}. Because the simulations serve as training data for data-driven surrogate models, we prioritize scalable dataset generation and consistent geometric representations while retaining the moving heat source model. 

\subsection*{Surrogates for AM Thermal, Stress, and Distortion Fields}
Due to the high computational cost of thermo-mechanical simulations of additive manufacturing processes, surrogate modeling has been widely explored to accelerate the prediction of quantities such as temperature histories, residual stresses, and part distortions. These approaches include reduced-order models, statistical surrogates, and data-driven machine learning methods that approximate mappings between process parameters, geometry descriptors, and simulation outputs \cite{xia2025surrogate, chaudhry2025data, tian2025real, dong2022part, liao2023hybrid, yaseen2023fast}. Within this broader landscape, data-driven methods for predicting spatially resolved physical fields from simulation data have received increasing attention, including convolutional neural networks, dimensionality-reduction-based surrogates, operator learning, and related architectures that map geometry and process parameters to temperature, displacement, or stress distributions \cite{dong2022part, xia2025surrogate, chaudhry2025data, tian2025real}.


In this work, we focus on making accurate and fast predictions of post-cooling displacement fields for DED with variable geometry and process conditions. Our objective is not simply to model parameter-to-field mappings, but to do so in a way that remains data efficient while supporting controllable generation across design and manufacturing parameters. Our hypothesis is that a linear affine reconstruction mechanism in the neural implicit space can provide a combination of accuracy, control, and data efficiency, even when the output is a complex spatial field and the number of training data is limited.

\subsection*{Neural operators, Neural Fields, and Field Prediction}
Because our goal is to predict entire spatial fields, it is important to consider representation methods that are well suited to continuous field modeling rather than only low-dimensional summary prediction or singular performance metrics. Neural operators and neural implicit fields are particularly relevant in this context because they provide ways to represent and learn continuous functions over space while accommodating parametric variations across samples \cite{kovachki2023neural, xie2022neural}.

Implicit neural representations model spatial signals using coordinate-based neural networks. Techniques such as Fourier feature mappings and sinusoidal activation functions mitigate spectral bias in multilayer perceptrons and enable accurate representation of high-frequency spatial variation \cite{tancik2020fourier, mildenhall2021nerf, sitzmann2019siren}. Conditioning these representations on external parameters is typically achieved by concatenating parameters to spatial coordinates or injecting them into intermediate layers \cite{perez2018film, ha2016hypernetworks}.

In parallel, neural operators provide a framework for learning mappings between function spaces and have recently emerged as powerful tools for modeling parametric families of partial differential equation solutions \cite{wu2024transolver, kovachki2023neural, li2020fourier}. For example, DeepONet introduces a branch–trunk architecture capable of learning nonlinear operators from scattered observations \cite{lu2021learning}. Such approaches illustrate the potential of operator-learning methods for accelerating physics-based simulations, which is included as a baseline in this work.



Recent work has also explored performing generation directly in neural network weight space. In particular, LAMP (Linear Affine Mixing of Parametric shapes) proposes a data-efficient framework for controllable 3D geometry generation by aligning exemplar-specific signed distance function decoders through overfitting from a shared initialization and synthesizing new geometries via affine mixing of their weights under parameter constraints \cite{nehme2025lamp}. This approach demonstrates that when networks are aligned within a common weight-space basin, linear combinations of weights can produce coherent and controllable outputs while requiring relatively few training samples.

Our approach builds on coordinate-based field representations but differs in the learning and inference methodology. Instead of learning a direct parameter-and-coordinate-to-field mapping, FLARE reconstructs network weights as affine combinations of models from the training set. This design is motivated by reconstruction-based representations observed in geometry generation and manifold learning literature \cite{roweis2000nonlinear, nehme2025lamp}. In contrast to LAMP, which focuses on geometry generation using implicit surface decoders, we extend affine weight-space reconstruction to implicit neural fields representing physical quantities, enabling parameter-controlled prediction of spatial simulation fields.

\section{Dataset Generation}



\subsection*{Modeling of the DED Process}


This section describes the thermo-mechanical finite element model we developed to simulate the DED process. The model simulates the displacement, stress, and strain (elastic, plastic, and thermal) throughout the manufacturing process. Instead of using the geometry-free approach with a temperature-based activation criterion proposed by Yushu et al. \cite{yushu2022directed}, we adopt a predefined geometry with a geometric activation criterion to reduce memory usage and ensure uniform geometric output, similar to the approaches in \cite{stender2018thermal, yang2016finite,heigel2015thermo}. We simulate the full deposition phase followed by a cooling phase, during which the laser power is set to zero and the part cools toward ambient temperature. We then use the post-cooling in-situ displacement field of the printed part for surrogate modeling.

We use the post-cooling in-situ displacement field for surrogate modeling because it represents the final as-built state after thermal transients have dissipated. This field is directly relevant to dimensional accuracy, geometric compensation, and manufacturing decision-making, since it reflects the residual deformation that remains in the part at the end of the process.

\subsubsection*{Material Deposition Model}
\begin{figure}
    \centering
    \includegraphics[width=1\linewidth]{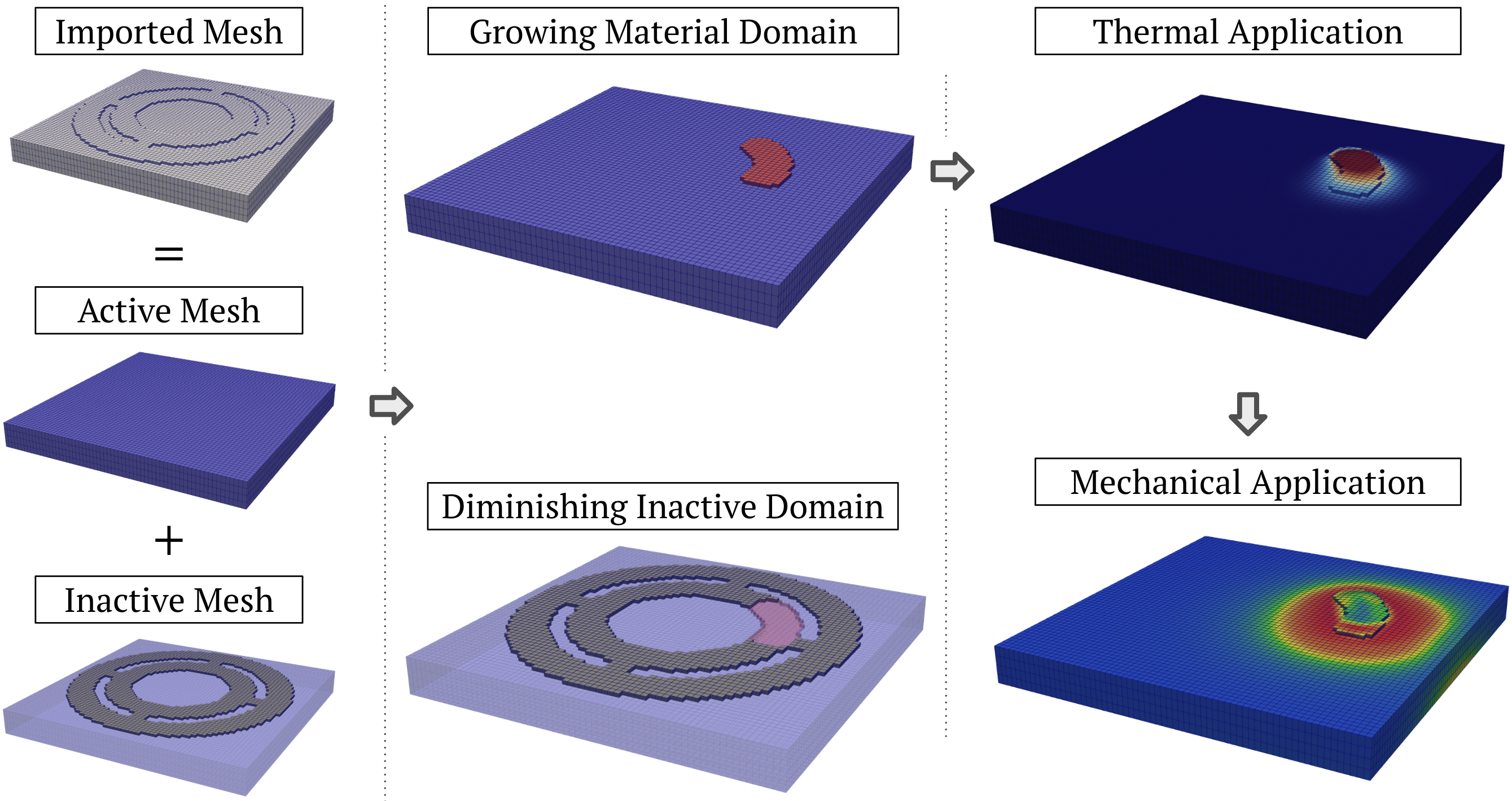}
    \caption{DED Modeling Schematic}
    \label{fig:DEDModelingScheme}
\end{figure}

The analysis employs a one-way coupled thermal-mechanical workflow using MOOSE's multi-application (multi-app) architecture, wherein the resulting temperature field from the thermal problem is transferred to the mechanical problem at each time step \cite{gaston2015physics, giudicelli2025data}. The thermal and mechanical analyses share an identical mesh comprised of the active domain ($\Omega_a$) and inactive domain ($\Omega_i$). The active domain is then subdivided into the deposited material domain ($\Omega_m$) and substrate domain ($\Omega_s$) such that $\Omega_a=\Omega_m \cup \Omega_s$. As the mesh is imported into MOOSE, the printable geometry and substrate are assigned to $\Omega_i$ and $\Omega_a=\Omega_s$, respectively (Fig. \ref{fig:DEDModelingScheme}). At this stage, the active deposited material domain, $\Omega_m$, is effectively empty. The substrate subdomain does not evolve throughout the analysis.

At each time step, an imaginary laser is moved across the inactive domain, $\Omega_i$. Then any element inside this cylinder is activated and transferred to the active domain, $\Omega_a$. Element activation is governed by a proximity-based criterion that determines whether an element lies within the effective deposition zone of the laser at the current time step. Specifically, for each element in the inactive subdomain, the horizontal distance from the element to the current laser position is evaluated against a time-varying activation radius. If any Gaussian integration point of the element satisfies this spatial criterion and the element's vertical coordinate does not exceed the current deposition layer height, the element is flagged for activation. The activation radius and deposition height are prescribed as piecewise functions. As time progresses, $\Omega_a$ grows, while $\Omega_s$ shrinks.

\subsubsection*{Thermal Model}
After elements are activated, the thermal application solves the transient heat conduction equation over the active domain, including both the substrate and deposited material. We obtain the temperature field by solving the transient heat equation:
\begin{equation}
\rho\, c(\theta)\, \frac{\partial \theta}{\partial t} = \nabla \cdot \left[\kappa(\theta)\,\nabla\theta\right] + Q(\mathbf{x}, t) \quad \text{in } \Omega_a
\label{eq:thermalEqn}
\end{equation}
where $\theta$ is the temperature, $\rho$ is the material density, $c(\theta)$ is the temperature-dependent specific heat, $\kappa(\theta)$ is the temperature-dependent thermal conductivity, and $Q(\mathbf{x}, t)$ is the volumetric heat source representing the laser energy input. We model the laser as a Gaussian heat source \cite{denlinger2017thermomechanical, michaleris2014modeling, irwin2016line, luo2019numerical}.
To avoid numerical inaccuracies that arise when the laser traverses more than one element per time step, we employ a hybrid heat source model that switches between the Gaussian point heat source and a time-averaged Gaussian line heat source \cite{irwin2016line, yushu2022directed}.
When the distance traveled by the laser within a single time step exceeds a threshold length or time, the line-averaged formulation is used; otherwise, the point-source formulation is applied.

A Dirichlet boundary condition is prescribed at the bottom surface of the substrate, $\Gamma^{z^-}_s$:
\begin{equation}
    \Gamma^{z^-}_s = \left\{ \boldsymbol{x} \in \partial \Omega_s \;\middle|\; \hat{\boldsymbol{n}}(\boldsymbol{x}) \cdot \boldsymbol{e}_z < 0 \right\}
\end{equation}
where $\hat{\boldsymbol{n}}(\boldsymbol{x})$ is the unit normal at point $\boldsymbol{x}$ on the surface $\partial\Omega$ and $\boldsymbol{e}_z$ is the unit vector in the $z$ direction.

The treatment of convective and radiative boundary conditions in additive manufacturing simulations is a complex and actively studied topic \cite{heigel2015thermo, ning2019analytical, li2019estimation}. Given the challenges associated with accurately characterizing environmental heat transfer, this dataset adopts a simplified treatment in which all free surfaces not prescribed by a Dirichlet condition exchange heat with the environment through convection at the ambient temperature. The aggregated boundary conditions are as follows: 

\begin{equation}
    \begin{cases}
        \theta = \bar{\theta}_{\text{substrate}}\; \text{on }\Gamma_s^{z^-}\\
        -\kappa(\theta)\ \nabla\theta \cdot\boldsymbol{\hat{n}} = h(\theta)\cdot\left[\theta-\bar{\theta}_\infty\right]\; \text{on }\partial\Omega_a\setminus\Gamma_s^{z^-}
    \end{cases}
\end{equation}
where $h(\theta)$ is the heat transfer coefficient and $\bar{\theta}_\infty$ is the ambient temperature. 

\subsubsection*{Mechanical Model}
After the thermal response is acquired, the temperature field is transferred to the mechanical application. We adopt the formulation outlined by Yushu et al. \cite{yushu2022directed}. In the mechanical application, we solve the quasi-static mechanical equilibrium equation (Eq. \ref{eq:momentum}) over the active domain, $\Omega_a$.
\begin{equation}
    \nabla \cdot \boldsymbol{\sigma} + \mathbf{b} = \mathbf{0} 
    \quad \text{in } \Omega_a = \Omega_s \cup \Omega_m
    \label{eq:momentum}
\end{equation}
where $\boldsymbol{\sigma}$ is the Cauchy stress tensor and $\mathbf{b}$ is the body force vector. The model employs the incremental infinitesimal strain formulation. The total strain ($\boldsymbol{\varepsilon}$) tensor is the symmetric part of the displacement gradient tensor and additively decomposed into elastic, plastic, and thermal components:
\begin{align}
    \boldsymbol{\varepsilon} &= \dfrac{1}{2}\left(\nabla \boldsymbol{u} + \left(\nabla \boldsymbol{u}\right)^\top\right)\\
    \boldsymbol{\varepsilon} &= \boldsymbol{\varepsilon}_e
                             + \boldsymbol{\varepsilon}_p
                             + \boldsymbol{\varepsilon}_\theta
    \label{eq:strain_decomp}
\end{align}
The elastic behavior is governed by a temperature-dependent isotropic elasticity 
tensor, constructed from the Young's modulus $E(\theta)$ and Poisson's ratio 
$\nu(\theta)$:
\begin{equation}
    \boldsymbol{\sigma} = \mathbb{C}(\theta) : \boldsymbol{\varepsilon}_e
    \label{eq:elastic}
\end{equation}
where $\mathbb{C}(\theta)$ is the fourth-order elasticity tensor. We apply the temperature dependence of the elastic properties through piecewise linear interpolation of tabulated values.


We adopt a von Mises plasticity model with isotropic hardening. The strength coefficient, $K$ and strain hardening exponent, $n$, is tabulated in Table \ref{tab:const_props} 
\cite{yushu2022directed}.

We compute thermal strains assuming isotropic thermal expansion:
\begin{equation}
    \boldsymbol{\varepsilon}^\theta = \beta\,(\theta - \theta_0)\,\mathbf{I}
    \label{eq:thermal_strain}
\end{equation}
where $\beta$ is the coefficient of thermal expansion, $\mathbf{I}$ is the second-order identity tensor, and $\theta_0$ is the stress-free reference temperature. Notably, the value of $\theta_0$ differs between the substrate and the deposited material. For the substrate, $\theta_0$ is set to room temperature, reflecting its initial stress-free state prior to deposition. For the deposited material, $\theta_0$ is set to the melting temperature, since each newly activated element is assumed to be stress-free upon activation. This choice is consistent with the formulation of Yushu et~al. and offers numerical stability advantages by minimizing thermal-induced stresses at the instant of activation \cite{yushu2022directed}.

The only mechanical boundary condition is of Dirichlet type at the bottom face of the substrate:

\begin{equation}
    \boldsymbol{u} = \boldsymbol{0}\; \text{on }\Gamma_s^{z^-}
\end{equation}



\subsection*{Geometry Generation and Meshing}
\begin{figure}
    \centering
    \includegraphics[width=1\linewidth]{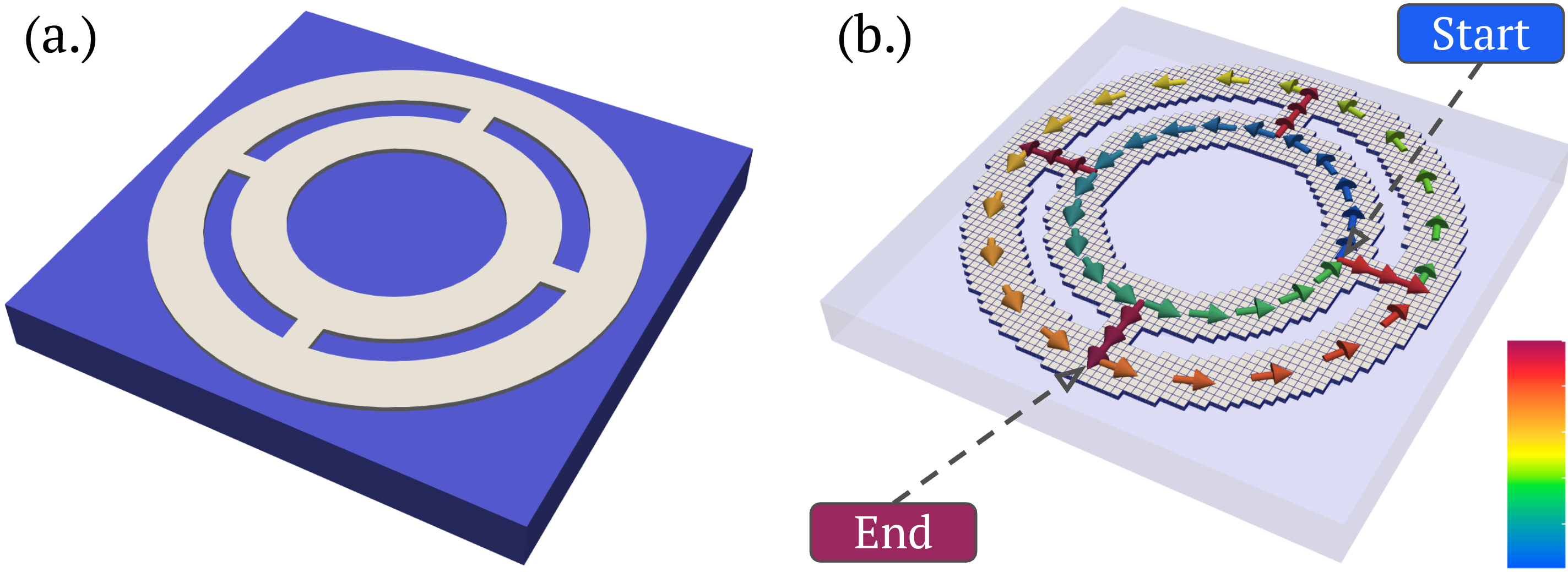}
    \caption{Geometry, Mesh, and Printing Path}
    \label{fig:geometryMeshPath}
\end{figure}

Our industry collaborators provided the geometry used in this study based on their internal use case, shown in Fig. \ref{fig:geometryMeshPath}a. The geometry consists of two concentric rings with different thicknesses. This configuration serves as a proof-of-concept geometry for evaluating the proposed surrogate modeling framework while maintaining geometric variability.

We created the CAD geometry with CadQuery \cite{cadquery_contributors_2025_14590990}. To generate the finite element mesh, the workflow employs a voxelization-based meshing procedure. First, we construct a structured grid over the bounding volume of the CAD model. Each grid element is then evaluated to determine whether its centroid lies within the solid geometry. Elements satisfying this criterion are retained, yielding a structured voxel mesh.

A sample mesh is illustrated in Fig. \ref{fig:geometryMeshPath}b. We selected the mesh resolution to balance computational cost with dataset generation requirements. Specifically, we chose the element size such that a single simulation could be completed in no more than 3 hours, running on 16 CPU cores of a local workstation. On average, each model has 32k nodes, and each data point has approximately 500 solved timesteps. 

All samples follow the same sequential print path parameterization shown in Fig. \ref{fig:geometryMeshPath}b. For consistency across all samples, the structures are fabricated as single-layer deposits following this predefined path.

\subsection*{Parameter Sampling}

The parameter ranges used to generate the dataset are summarized in Table \ref{tab:param_ranges}. We generated 100 parameter configurations using Latin hypercube sampling (LHS) across the seven-dimensional parameter space to ensure broad and uniform coverage of the design domain. We selected geometric parameters such that the extreme values do not lead to geometric overlap between the inner and outer rings, thereby ensuring physically valid geometries throughout the sampled space.


We chose manufacturing parameters informed by values reported in the literature for directed energy deposition of stainless steel. The layer thickness (height) and deposition velocity were selected within ranges consistent with typical process settings \cite{davoodi2022additively, prasad2020powder, sayyar2023directed, pereira2023semi, li2021experimental, ronda2022influence, kim2024microstructural, tepponen2026mechanical, ziqiang2025investigation, elgazzar2025cost}. The laser power range was intentionally extended beyond values commonly reported for stainless steel additive manufacturing to ensure that approximately half of the sampled parameter combinations are deemed infeasible under our feasibility criteria.

\begin{table}[H]
\centering
\caption{Parameter ranges used for dataset generation.}
\label{tab:param_ranges}
\begin{tabular}{llcl}
\toprule
\textbf{Parameter} & \textbf{Symbol} & \textbf{Range} & \textbf{Units} \\
\midrule
Outer ring center radius     & $r_{\text{out}}$ & 35 -- 40 & mm \\
Outer ring thickness         & $t_{\text{out}}$ & 5 -- 10  & mm \\
Inner ring center radius     & $r_{\text{in}}$  & 20 -- 25 & mm \\
Inner ring thickness         & $t_{\text{in}}$  & 5 -- 10  & mm \\
Height                 & $h$              & 0.2 -- 1.0 & mm \\
Laser power                  & $P$              & 5 -- 10  & kW \\
Deposition velocity          & $v$              & 5 -- 10  & mm/s \\
\bottomrule
\end{tabular}
\end{table}

The inclusion of both feasible and infeasible parameter combinations is by design. In geometry-driven element activation approaches, elements are activated according to a geometric criterion rather than melt pool physics. As a result, not all activated elements correspond to physically realizable melt conditions. Including such cases allows the dataset to capture both valid and invalid processing regimes. While not pursued here, this distinction can provide valuable guidance for developing generative models capable of proposing feasible parameter combinations for a given target geometry \cite{regenwetter2024constraining}.

Feasibility is determined by analyzing the temperature history of nodes within the deposited material. A parameter set is classified as feasible if more than 99\% of the nodes in the deposited domain experience temperatures exceeding the material melting temperature. The 99\% threshold is used instead of a strict 100\% requirement because a small number of seed elements are introduced to facilitate initialization of the active–inactive element domains.

\begin{figure}
    \centering
    \includegraphics[width=1\linewidth]{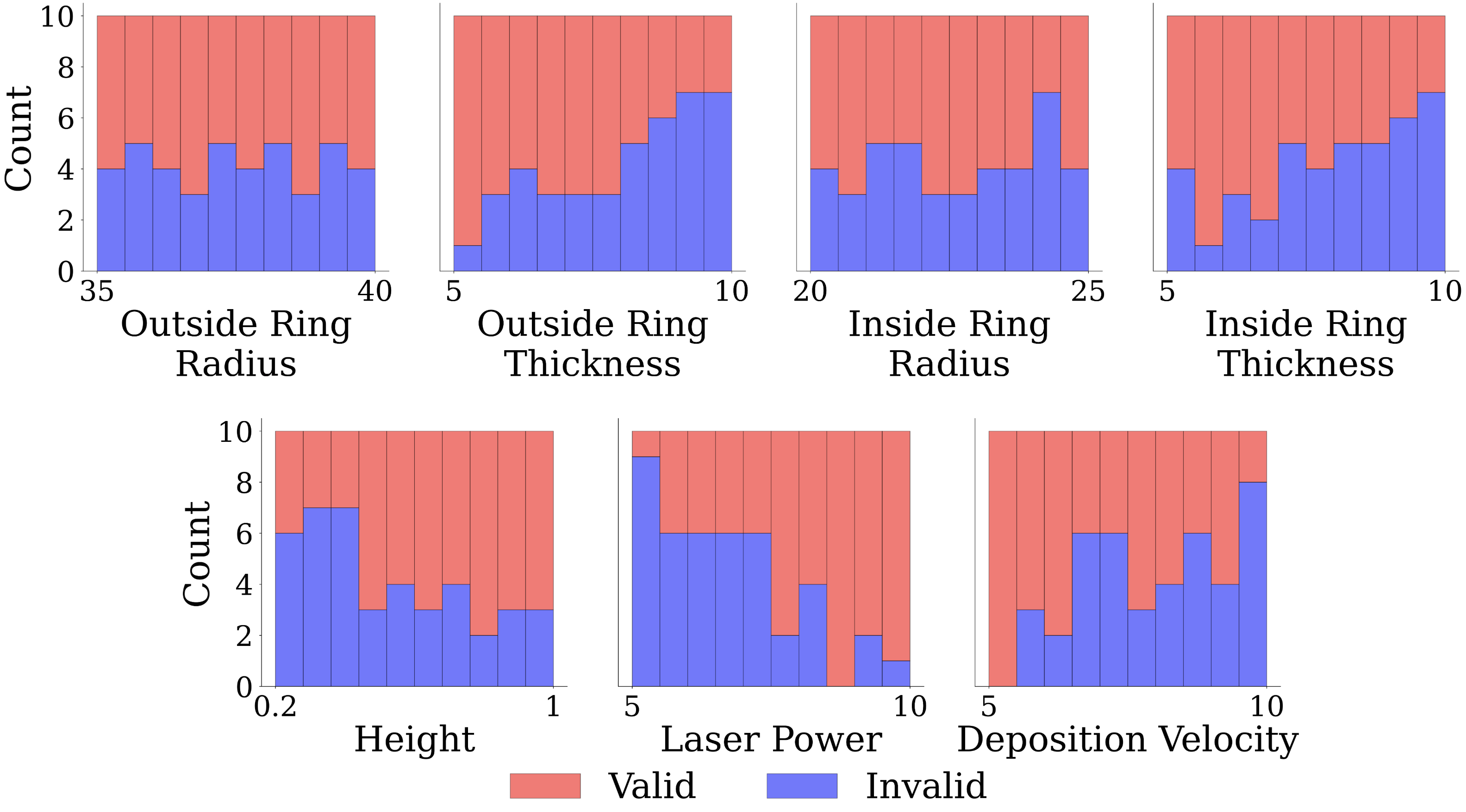}
    \caption{Sampled Parameters and Feasibility Distribution}
    \label{fig:datasetSampling}
\end{figure}

Figure \ref{fig:datasetSampling} displays the distribution of the sampled parameters. Red denotes the valid parameter distribution, while blue denotes invalid parameter distribution. 

\section{Method}
\label{sec:method}

\begin{figure*}
    \centering
    \includegraphics[width=1\linewidth]{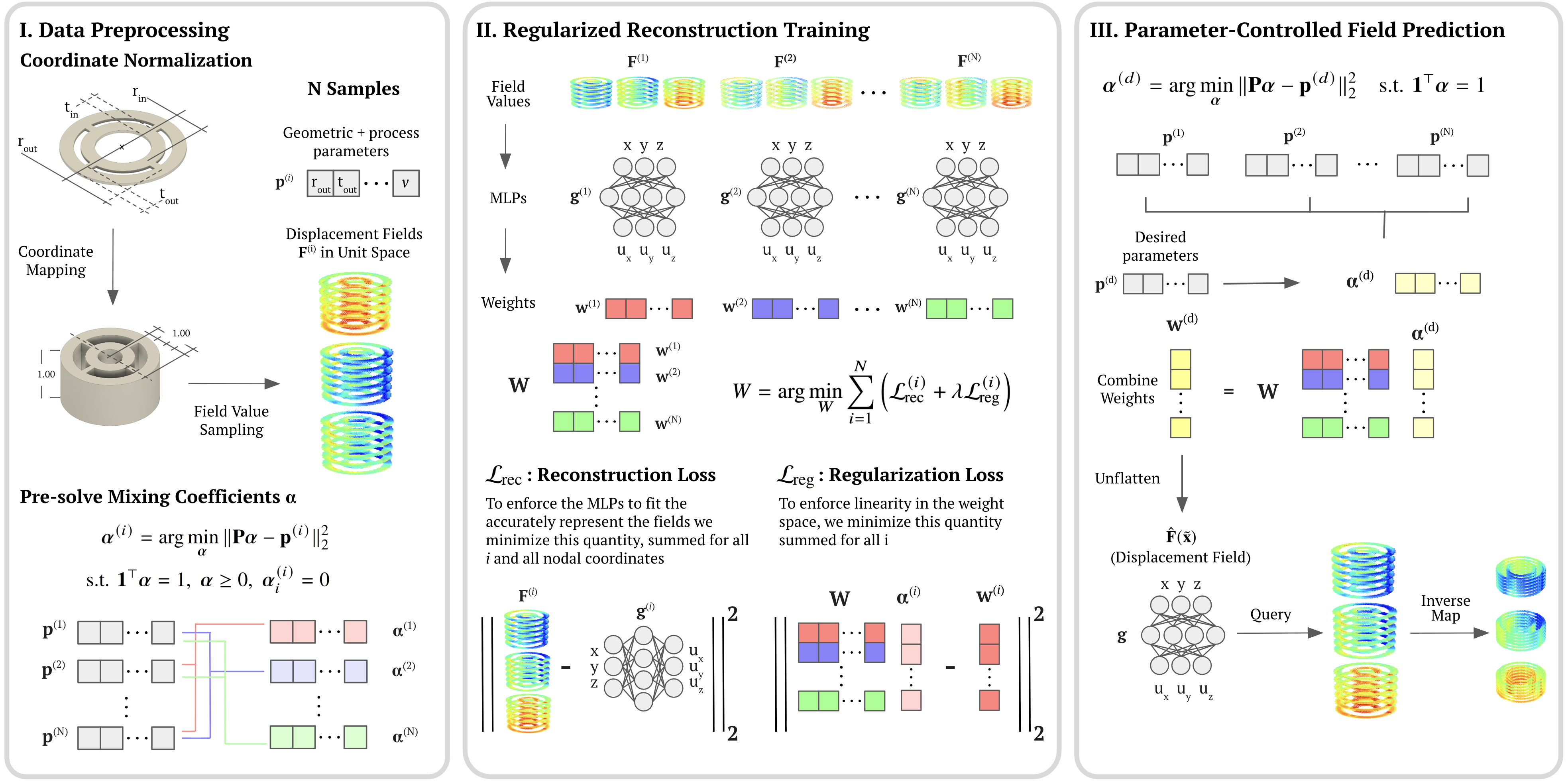}
    \caption{Overview of the FLARE method. Each physical field is represented by an implicit neural network. A linear relationship between process parameters and neural network weights is enforced through affine reconstruction regularization, enabling controllable field generation.}
    \label{fig:mainArchitectureFigure}
\end{figure*}

We introduce \textbf{FLARE} (\textbf{F}ields via \textbf{L}inear \textbf{A}ffine \textbf{R}econstruction in w\textbf{E}ight-space), a framework for predicting physical fields in low-data regimes by constructing a parameter-controllable latent representation in neural network weight space (Fig. \ref{fig:mainArchitectureFigure}).

The key idea is to represent each field as a neural network while enforcing a linear relationship between the parameters and the network weights. This produces a structured latent space where new fields can be generated by solving a small optimization problem in parameter space and reconstructing the corresponding neural network weights.

Our approach builds on LAMP’s affine weight-space mixing framework, where exemplar networks are aligned through overfitting from a shared initialization and new instances are generated via parameter-constrained affine combinations of their weights \cite{nehme2025lamp}. We extend this principle to implicit neural fields for controllable physical field prediction.

\subsection*{Problem Setup}

We consider $N$ geometries consisting of two concentric rings with varying geometric and process parameters. 

Each design is represented by a parameter vector

\begin{equation}
\mathbf{p}^{(i)} \in \mathbb{R}^k ,
\end{equation}

where $k=7$ corresponds to the parameters listed in Table \ref{tab:param_ranges}. Collectively,

\begin{equation}
\mathbf{P} =
\begin{bmatrix}
\mathbf{p}^{(1)} & \dots & \mathbf{p}^{(N)}
\end{bmatrix}
\end{equation}

For each design $i$, a simulation produces a field defined on a discretized geometry with $m^{(i)}$ nodal coordinates
\begin{equation}
\mathbf{x}_j^{(i)}=(x^{(i)}_{j},y^{(i)}_{j},z^{(i)}_{j}), \quad j=1, \dots, m^{(i)}
\end{equation}
and field values
\begin{equation}
\textbf{F}_j^{(i)}=\textbf{F}(\mathbf{x}_j^{(i)};\mathbf{p}^{(i)})
\end{equation}
Our goal is to construct a controllable generator that predicts the spatial field for some desired arbitrary parameters $\mathbf{p}^{(d)}$.

\subsection*{Coordinate Normalization}

To enable a shared neural representation across geometries with varying dimensions, we map nodal coordinates to a canonical domain. We rescale radial coordinates so that the boundaries of the concentric rings align across samples (Fig. \ref{fig:mainArchitectureFigure}).

Let

\begin{equation}
\tilde{\mathbf{x}}_j^{(i)} = T(\mathbf{x}_j^{(i)}, \mathbf{p}^{(i)})
\end{equation}

denote normalized coordinates obtained through a geometry-dependent mapping $T$. The mapping is invertible, allowing predicted fields to be evaluated in the original physical domain.

\subsection*{Implicit Field Representation}

Each field is represented by an implicit neural field

\begin{equation}
\textbf{g}^{(i)}(\tilde{\mathbf{x}};\mathbf{w}^{(i)}),
\end{equation}

where $\mathbf{w}^{(i)}$ denotes the flattened network weights (Fig. \ref{fig:mainArchitectureFigure}). The input coordinates are augmented with Fourier positional encoding:

\begin{equation}
\phi(\tilde{\mathbf{x}}) =
[\tilde{\mathbf{x}};\gamma(\tilde{\mathbf{x}})]
\end{equation}

The reconstruction loss for each sample is

\begin{equation}
\mathcal{L}_{\text{rec}}^{(i)} =
\frac{1}{m_i}\sum_{j=1}^{m_i}
\|\textbf{g}^{(i)}(\tilde{\mathbf{x}}_j^{(i)};\mathbf{w}^{(i)})-\textbf{F}_j^{(i)}\|_2^2
\end{equation}

To ensure that all networks occupy a consistent region of weight space, we first overfit a single sample using an MLP. The resulting weights $\mathbf{w}_0$ are then used to initialize all other networks. This initialization keeps the solutions close in weight space and facilitates learning a structured latent representation.

\subsection*{Affine Weight-Space Regularization}

To enable controllable field generation, FLARE enforces that neural network weights follow the same affine structure as the parameter space.

For each training sample $i$, we compute coefficients $\boldsymbol{\alpha}^{(i)}$ that reconstruct its parameters from the remaining training samples:

\begin{equation}
\begin{aligned}
\boldsymbol{\alpha}^{(i)}=
\arg\min_{\boldsymbol{\alpha}}
\|\mathbf{P}\boldsymbol{\alpha}-\mathbf{p}^{(i)}\|_2^2 \\
\text{s.t. } \mathbf{1}^\top\boldsymbol{\alpha}=1, \;
\boldsymbol{\alpha}\ge0, \; \boldsymbol{\alpha}^{(i)}_i=0 .
\end{aligned}
\end{equation}

These coefficients define the local affine structure of the parameter space. We transfer this structure to weight space by enforcing

\begin{equation}
\mathcal{L}_{\text{reg}}^{(i)}=
\|\mathbf{W}\boldsymbol{\alpha}^{(i)}-\mathbf{w}^{(i)}\|_2^2 ,
\end{equation}



where

\begin{equation}
\mathbf{W}=
\begin{bmatrix}
\mathbf{w}^{(1)} & \dots & \mathbf{w}^{(N)}
\end{bmatrix}
\end{equation}

stacks the network weights.

The total training objective is

\begin{equation}
\mathcal{L}=
\sum_{i=1}^N
\left(
\mathcal{L}_{\text{rec}}^{(i)}
+
\lambda
\mathcal{L}_{\text{reg}}^{(i)}
\right)
\end{equation}

This regularization encourages the weight space to form a linear latent manifold aligned with the parameter space.

\subsection*{Field Generation}

At inference, a desired parameter vector $\mathbf{p}^{(d)}$ is expressed as an affine combination of training parameters

\begin{equation}
\boldsymbol{\alpha}^{(d)} =
\arg\min_{\boldsymbol{\alpha}}
\|\mathbf{P}\boldsymbol{\alpha}-\mathbf{p}^{(d)}\|_2^2
\quad
\text{s.t. } \mathbf{1}^\top\boldsymbol{\alpha}=1
\end{equation}

The corresponding neural network weights are reconstructed as

\begin{equation}
\mathbf{w}^{(d)} = \mathbf{W}\boldsymbol{\alpha}^{(d)}
\end{equation}

The predicted field is then evaluated as

\begin{equation}
\hat{\textbf{F}}(\tilde{\mathbf{x}})=\textbf{g}(\tilde{\mathbf{x}};\mathbf{w}^{(d)})
\end{equation}

Because inference only requires solving a small optimization problem in parameter space followed by a linear reconstruction of network weights, the method enables fast field generation. Implementation details such as network architectures, Fourier feature configuration, and training hyperparameters are provided in Appendix \ref{app:implementation}.


\section{Experiments, Results, and Discussion}
\label{sec:experiments}

We evaluate FLARE on the task of predicting displacement fields in the post-cooling stage. The inputs consist of geometric parameters and process parameters, and the outputs are spatial displacement fields defined over mesh nodes. We consider both in-distribution and extrapolation settings and compare against several parameter-conditioned baselines.

\subsubsection*{Experimental Setup}
Each simulation produces displacement fields $\{u_x,\ u_y,\ u_z\}$ defined at the nodes of the discretized geometry. The input parameters include geometric parameters of the two-ring structure together with process parameters controlling the printing process (laser power and deposition velocity).

\begin{figure}[H]
    \centering
    \includegraphics[width=1\linewidth, scale=0.50]{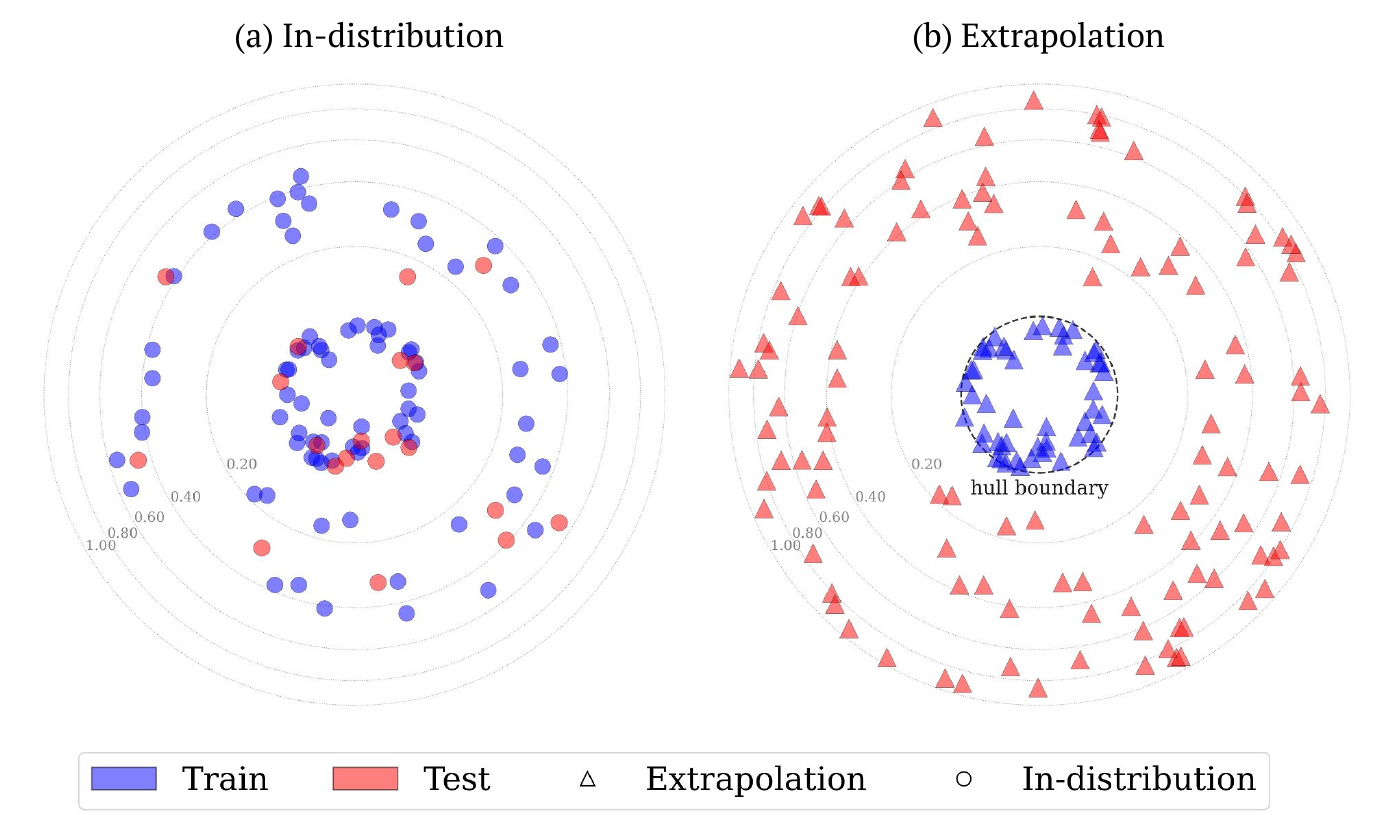}
    \caption{Radial hull-distance visualization of the parameter space showing the train and test sets for the interpolation and extrapolation experiments}
    \label{fig:MDS}
\end{figure}

\begin{figure*}
    \centering
    \includegraphics[width=1\linewidth]{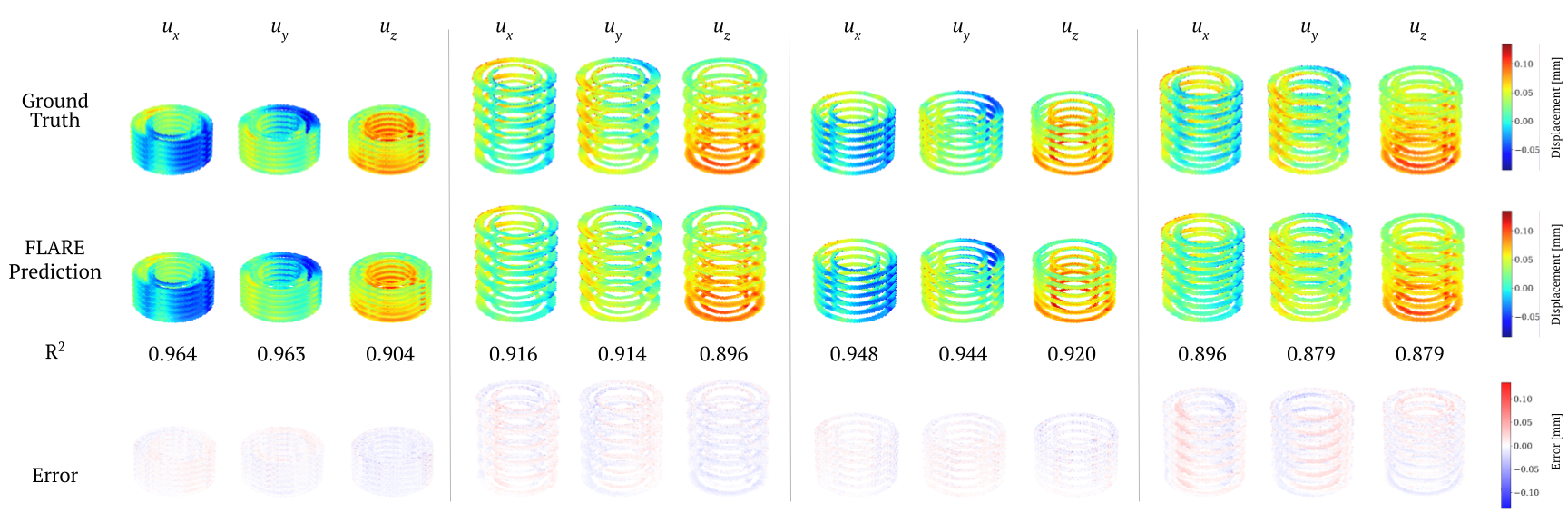}
        \caption{Sample Ground Truth, Prediction, and Error}
    \label{fig:predGT}
\end{figure*}

\begin{table*}[t]
\centering
\caption{In-Distribution performance}
\label{tab:interp_results}
\small
\setlength{\tabcolsep}{5pt}

\begin{tabular}{lcccccccccccc}
\toprule
& \multicolumn{3}{c}{$R^2\uparrow$} 
& \multicolumn{3}{c}{RMSE ($\times 10^{-3}$)$\downarrow$}
& \multicolumn{3}{c}{Weighted $R^2\uparrow$} 
& \multicolumn{3}{c}{Weighted RMSE ($\times 10^{-3}$)$\downarrow$} \\
\cmidrule(lr){2-4} \cmidrule(lr){5-7} \cmidrule(lr){8-10} \cmidrule(lr){11-13}
Method 
& $u_x$ & $u_y$ & $u_z$
& $u_x$ & $u_y$ & $u_z$
& $u_x$ & $u_y$ & $u_z$
& $u_x$ & $u_y$ & $u_z$ \\
\midrule

Nearest Neighbor 
& 0.8416 & 0.8309 & 0.6779 & 11.37 & 11.29 & 12.82 
& 0.8940 & 0.8982 & 0.5467 & 13.67 & 13.31 & 13.07 \\

MLP Concatenate  
& 0.9345 & 0.9295 & 0.9024 & 7.647 & 7.646 & 7.973 
& 0.9595 & 0.9595 & 0.9002 & 9.381 & 9.308 & 6.814 \\

MLP FiLM         
& 0.9391 & 0.9352 & 0.9055 & 7.326 & 7.292 & 7.802 
& 0.9614 & 0.9602 & 0.8976 & 9.023 & 9.077 & 6.820 \\

DeepONet         
& 0.9263 & 0.9202 & 0.8763 & 8.077 & 8.123 & 8.924 
& 0.9570 & 0.9577 & 0.8660 & 9.615 & 9.510 & 7.829 \\

LAMP             
& 0.8966 & 0.8945 & 0.8389 & 9.598 & 9.336 & 10.13 
& 0.9427 & 0.9409 & 0.8340 & 11.14 & 11.16 & 8.789 \\

FLARE (ours)     
& \textbf{0.9483} & \textbf{0.9454} & \textbf{0.9201} 
& \textbf{6.754} & \textbf{6.701} & \textbf{7.182} 
& \textbf{0.9715} & \textbf{0.9719} & \textbf{0.9168} 
& \textbf{7.822} & \textbf{7.724} & \textbf{6.167} \\

\bottomrule
\end{tabular}
\end{table*}

We evaluate the sensitivity of FLARE across train set sizes and test two generalization regimes. To test \textbf{sensitivity}, we start with a random 80/20 train/test split. Holding the test split constant, the training set is selected by maximizing the minimum distance to any existing point, starting with the centroid of the normalized parameter space. Using this greedy algorithm, we create train sets of sizes 10, 20, 40, 60, and 80. 

In the \textbf{in-distribution} setting, we randomly partition the 100 Latin hypercube samples into an 80-sample training set and a 20-sample test set (Fig. \ref{fig:MDS}a).


In the \textbf{extrapolation} setting, we train on 51 basis experiments and evaluate on 113 test points. The test set includes 49 trimmed experiments outside the selected basis and 64 additional experiments sampled from the corner points of the parameter bounds (Fig.~\ref{fig:MDS}b), assessing generalization beyond the observed region. The basis set is constructed by uniformly and symmetrically trimming the parameter space to retain $\approx$ 50 points within a 7D cube. The additional 64 test points are randomly selected from the 128 corner points of the corresponding 7D hypercube to probe extreme extrapolation.

\begin{table*}[t]
\centering
\caption{Extrapolation performance}
\label{tab:extrap_results}
\small
\setlength{\tabcolsep}{5pt}

\begin{tabular}{lcccccccccccc}
\toprule
& \multicolumn{3}{c}{$R^2\uparrow$} 
& \multicolumn{3}{c}{RMSE ($\times 10^{-3}$)$\downarrow$}
& \multicolumn{3}{c}{Weighted $R^2\uparrow$} 
& \multicolumn{3}{c}{Weighted RMSE ($\times 10^{-3}$)$\downarrow$} \\
\cmidrule(lr){2-4} \cmidrule(lr){5-7} \cmidrule(lr){8-10} \cmidrule(lr){11-13}
Method 
& $u_x$ & $u_y$ & $u_z$
& $u_x$ & $u_y$ & $u_z$
& $u_x$ & $u_y$ & $u_z$
& $u_x$ & $u_y$ & $u_z$ \\
\midrule

Nearest Neighbor 
& 0.7967 & 0.7902 & 0.6463 & 12.43 & 12.32 & 13.61 
& 0.8498 & 0.8542 & 0.4831 & 16.28 & 15.98 & 14.27 \\

MLP Concatenate  
& 0.8884 & 0.8857 & 0.8416 & 9.795 & 9.715 & 10.42 
& 0.9256 & 0.9264 & 0.8275 & 13.27 & 13.16 & 9.614 \\

MLP FiLM         
& 0.8476 & 0.8459 & 0.7220 & 11.49 & 11.32 & 13.12 
& 0.8788 & 0.8812 & 0.6576 & 16.67 & 16.44 & 13.09 \\

DeepONet         
& 0.8762 & 0.8656 & 0.8101 & 10.12 & 10.30 & 11.13 
& 0.9321 & 0.9289 & 0.7970 & 12.36 & 12.55 & 10.24 \\

LAMP             
& 0.7467 & 0.7351 & 0.6413 & 14.25 & 14.26 & 14.99 
& 0.8779 & 0.8773 & 0.6448 & 16.48 & 16.36 & 13.87 \\

FLARE (ours)     
& \textbf{0.8911} & \textbf{0.8887} & \textbf{0.8556} 
& \textbf{9.289} & \textbf{9.217} & \textbf{9.544} 
& \textbf{0.9404} & \textbf{0.9403} & \textbf{0.8458} 
& \textbf{11.27} & \textbf{11.23} & \textbf{8.748} \\

\bottomrule
\end{tabular}
\end{table*}

\subsubsection*{Baselines}
We compare FLARE against the nearest neighbor in parameter space, several parameter-conditioned neural field models, and operator-learning:
\begin{itemize}
\item \textbf{Nearest Neighbor}: the most similar data in the training set, evaluated using cosine similarity in the parameter space.
\item \textbf{LAMP} \cite{nehme2025lamp}: affine interpolation in weight space without the proposed regularization.
\item \textbf{MLP Concatenate}: an MLP that directly takes spatial coordinates concatenated with the parameter vector.
\item \textbf{MLP FiLM} \cite{perez2018film}: an MLP where intermediate activations are modulated using FiLM conditioning based on the parameter vector.
\item \textbf{DeepONet} \cite{lu2021learning}: a neural operator architecture for mapping parameter functions to spatial fields.
\end{itemize}

\subsubsection*{Evaluation Metrics}
Performance is measured using the coefficient of determination ($R^2$) and root mean squared error (RMSE). Additionally, a weighted variant of both metrics are considered to account for the bias of near-zero field values. The weighted metrics are defined as follows: 


\begin{align*}
R^2_{w^2} &= 1 -
\frac{\sum_{i=1}^{N} w_i (y_i^{\mathrm{true}}-\hat y_i)^2}
{\sum_{i=1}^{N} w_i (y_i^{\mathrm{true}}-\sum_{k=1}^{N} w_k y_k^{\mathrm{true}})^2} \\
\mathrm{RMSE}_{w^2} &= \sqrt{\sum_{i=1}^{N} w_i (y_i^{\mathrm{true}}-\hat y_i)^2} \text{, with } w_i = \frac{(y_i^{\mathrm{true}})^2}{\sum_{j=1}^{N}(y_j^{\mathrm{true}})^2}
\end{align*}


\subsubsection*{Train-set Size Sensitivity}
\begin{figure}[H]
    \centering
    \includegraphics[width=1\linewidth]{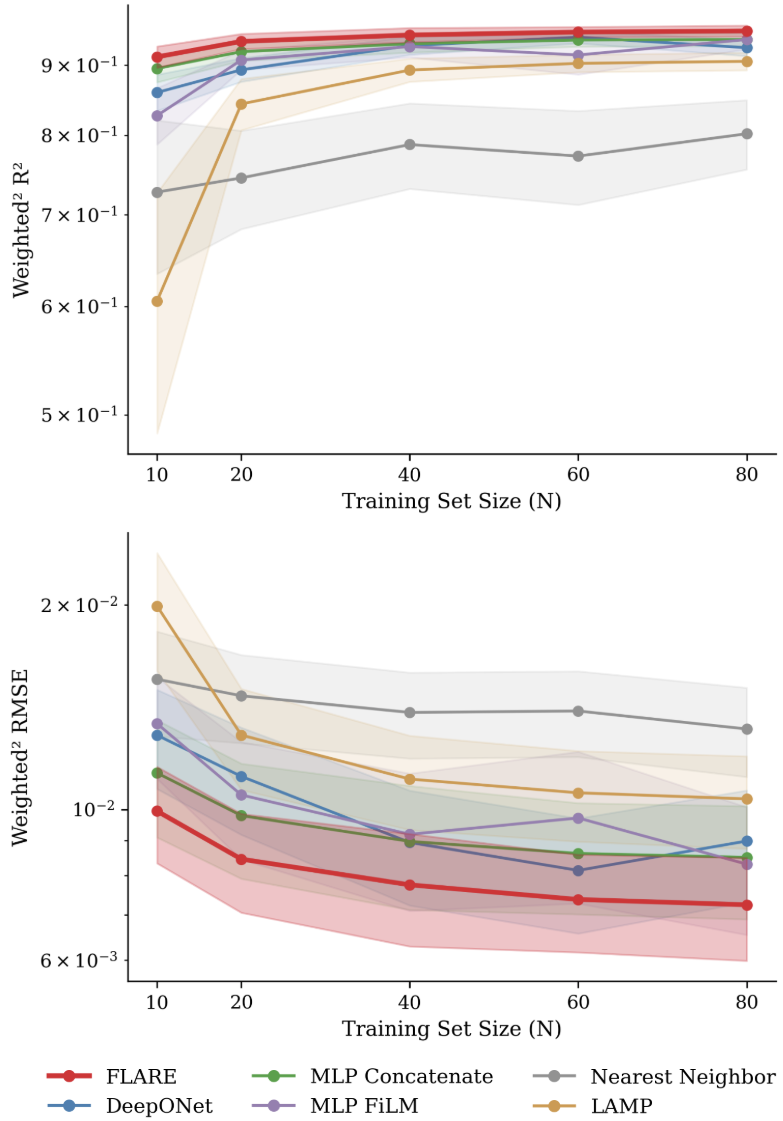}
    \caption{Metric vs Size of Training Set}
    \label{fig:trainSensitivity}
\end{figure}

Figure \ref{fig:trainSensitivity} reports results for the sensitivity of each model against a diminishing training set. FLARE consistently outperforms the baselines across all metrics. In this experiment, FLARE was consistently retrained for each test set to avoid data leakage from the precomputed parameter combination weights. Bands are only 0.5 standard deviation and represent the distribution of the metrics across the test samples.

\subsubsection*{In-distribution Results}
Table~\ref{tab:interp_results} reports results for the in-distribution setting. FLARE consistently outperforms the baselines across all metrics and displacement components. In general, prediction accuracy is highest for $u_x$ and $u_y$, while $u_z$ is more challenging to predict.



\subsubsection*{Extrapolation Results}
Table~\ref{tab:extrap_results} shows results for extrapolation experiments where test parameters lie outside the convex hull of the training set. 
FLARE maintains superior performance across all metrics compared to the baselines, highlighting the advantage of the affine weight-space reconstruction framework for controlled extrapolation.



\subsubsection*{Feasibility Prediction}
While field prediction is performed on the full dataset, the simulation procedure returns an output for every input configuration, so we also train a feasibility discriminator to determine whether a given combination of geometry and process parameters leads to a valid simulation based on our feasibility criteria. 

The dataset is moderately balanced, with 58\% valid and 42\% invalid samples. The trained discriminator using logistic regression with polynomial features and lasso regularization achieves 80\% accuracy and area under the Receiver Operating Characteristic (ROC) curve of 0.896 on a random 80/20 train/test split.

This component allows the framework to filter out parameter combinations that would lead to nonphysical conditions. The discriminator can be used either in conjunction with FLARE during inference or independently to guide generative models toward feasible regions of the parameter space.

\subsubsection*{Discussion}
Across both interpolation and extrapolation settings, FLARE consistently outperforms the baselines in predicting displacement fields. Furthermore, for identical architecture and MLP size, FLARE outperforms LAMP consistently in this application. This highlights the advantage of explicitly enforcing affine structure in weight space during training. We also observe that $u_z$ tends to exhibit lower prediction accuracy compared to $u_x$ and $u_y$, likely due to stronger nonlinearities in the vertical expansion and contraction induced during the printing process. 

Another notable observation is that we employ an inverse sampling method in training the MLPs, which consistently outperforms forward sampling. This highlights that while we require a unit-space mapping to exploit this method, we do not have to sample identically across that unit space. Overall, these results demonstrate that FLARE provides a robust and controllable framework for predicting manufacturing-induced displacement fields across a wide range of geometry and process parameters.

\subsubsection*{Computational Cost and Inference Speed}
The FLARE model trained on 80 samples required 1.25 hours of offline training on a single NVIDIA L40S. In contrast, generating a single additional data point with the finite element solver requires approximately 2.5 hours on 16 CPU cores. Evaluating FLARE for 20 samples on the same setup, without GPU acceleration, took 29 seconds in total, corresponding to approximately 1.45 seconds per sample. This yields an approximate 6200× reduction in per-query runtime relative to the finite element solver, while the one-time training cost is amortized over subsequent predictions.


\section{LIMITATIONS, FUTURE WORK AND CONCLUSION}
\label{sec:limitations}

Our coordinate normalization is tailored to the concentric-ring geometry studied here and may require adaptation for more complex shapes. In particular, the nonlinear coordinate mapping causes the spoke regions in our model to have a nonuniform sampling manifold in the unit space. This effect is illustrated in Fig.~\ref{fig:mainArchitectureFigure}. After the geometry is mapped into the unit space, the spokes are no longer parallel.

FLARE assumes that neural field weights vary approximately in an affine manner with respect to the parameter space. Although enforced during training, this assumption may limit performance for capturing highly nonlinear phenomena or parameters far from the training distribution. For instance, we observe lower accuracy for the vertical displacement component ($u_z$) across all tests, suggesting that the underlying behavior along the build direction is more difficult to capture than in the other two directions. Training also requires fitting a separate implicit field network per sample, which increases computational cost as the dataset grows. 

Future work will extend the framework to predict additional physical quantities, such as stress and strain fields, in addition to displacement. We also plan to expand the parameter space by incorporating additional manufacturing variables, including substrate temperature, laser radius, material properties, and printing path to better capture the full range of process variability in the directed energy deposition process.

The dataset used in this study was generated by sampling only seven parameters. This choice was made intentionally to limit the dimensionality of the parameter space and allow a clear demonstration of the proposed technique. While this simplified setting enables controlled evaluation of the method, practical additive manufacturing processes often involve a larger number of interacting parameters. Future work will investigate the scalability of the approach to higher-dimensional parameter spaces and more complex process conditions. 

Although the modeling approach is adapted from prior literature, experimental validation of the predicted field outputs lies beyond the scope of the present study. Accordingly, the results reported here focus on agreement within the simulation framework. 

This paper introduced FLARE, a data-efficient surrogate framework for predicting manufacturing-induced displacement fields using affine reconstruction in neural network weight space. By enforcing the structure of the implicit neural fields' weight space through regularization, the proposed approach enables accurate and controllable field generation under limited data. Experiments on thermo-mechanical DED simulations demonstrate that FLARE consistently outperforms parameter-conditioned neural field models and neural operator baselines in both interpolation and extrapolation regimes. These results highlight the potential of affine weight-space reconstruction as a robust strategy for surrogate modeling of physics-based manufacturing processes.



\section*{Acknowledgments}
The financial support of GE Vernova under a sponsored research agreement (MIT x GE Vernova Alliance) with the Massachusetts Institute of Technology is gratefully acknowledged.

\nocite{*}

\bibliographystyle{asmeconf}  
\bibliography{asmeconf-sample}
\newpage

\appendix

\section{Constants for simulation setup}
\label{apx:MatProp}
\begin{table}[H]
\centering
\caption{Constant inputs and material properties}
\label{tab:const_props}
\begin{tabular}{@{}llll@{}}
\toprule
\textbf{Parameter} & \textbf{Symbol} & \textbf{Value} & \textbf{Units} \\
\midrule
Density                          & $\rho$                   & $7.61\times10^{3}$ & \si{\kilogram\per\cubic\meter} \\
Thermal expansion coefficient    & $\beta$                  & $1.72\times10^{-5}$ & \si{\per\kelvin} \\
Substrate temperature            & $\bar{\theta}_{\text{substrate}}$    & 300                 & \si{\kelvin} \\
Ambient temperature              & $\bar{\theta}_\infty$    & 300                 & \si{\kelvin} \\
Melt temperature                 & $\theta_{\mathrm{melt}}$ & 1700                & \si{\kelvin} \\
Strength coefficient             & $K$                      & 847                 & --  \\
Strain hardening exponent        & $n$                      & 0.06                & --  \\
\bottomrule
\end{tabular}
\end{table}

\section{Implementation Details}
\label{app:implementation}

\subsection{Network Architecture}

Each field is represented by a multilayer perceptron (MLP). A separate network is trained for each training sample. The network maps spatial coordinates to displacement field values.

The input to the network consists of 3D spatial coordinates $(x,y,z)$ which are first transformed using Fourier positional encoding. We use $L=3$ frequency octaves, resulting in an encoded input dimension of:

\[
3 + 6L = 3 + 6 \times 3 = 21
\]

The encoded coordinates are processed by an MLP with four hidden layers of width 512 and ReLU activations. The network outputs a three-dimensional displacement vector:

\[
(u_x, u_y, u_z)
\]

\subsection{Input Structure}

Each training data point corresponds to a finite element simulation with a different set of manufacturing parameters. For each simulation, mesh nodes are inverse-sampled onto a shared unit representation outlined in Appendix \ref{apx: coordinate}. This produces a dataset consisting of

\begin{itemize}
\item \textbf{Inputs ($X$):} 3D coordinates of mesh nodes normalized to the unit domain $[-1,1]^3$. The same unit representation is used across all data points.
\item \textbf{Targets ($Y$):} displacement field values $(u_x,u_y,u_z)$ defined at those coordinates, which vary across simulations.
\end{itemize}

Each coordinate is expanded using Fourier positional encoding. For a point $(x,y,z)$ the encoding becomes

\begin{equation*}
\small
\begin{aligned}
(x,y,z) \rightarrow [&\,x,y,z, \\
&\sin(2^0x),\cos(2^0x),\sin(2^0y),\cos(2^0y),\sin(2^0z),\cos(2^0z),\\
&\sin(2^1x),\cos(2^1x),\sin(2^1y),\cos(2^1y),\sin(2^1z),\cos(2^1z),\\
&\sin(2^2x),\cos(2^2x),\sin(2^2y),\cos(2^2y),\sin(2^2z),\cos(2^2z)]
\end{aligned}
\end{equation*}

This produces a $21$-dimensional input vector per spatial point.

Each network therefore learns a mapping

\[
(x,y,z) \rightarrow (\text{u}_x,\text{u}_y,\text{u}_z)
\]

for a single simulation. Manufacturing parameters are not provided as network inputs; instead, parameter control is achieved through weight-space reconstruction using the affine mixing coefficients described in Section~\ref{sec:method}.

\subsection{Training Procedure}

Training is performed in two phases.

\textbf{Phase 1 (Base Training).}  
One random network is first trained independently to overfit its corresponding displacement field using mean squared error (MSE) reconstruction loss. This phase runs for $500{,}000$ epochs with early stopping.

\textbf{Phase 2 (Joint Training).}  
All networks are then jointly optimized for an additional $500{,}000$ epochs using the weights obtained from Phase 1 as initialization. During this phase both reconstruction loss and weight-space regularization are applied.

\subsection{Optimization}

Training uses the Adam optimizer. The learning rate is initialized at $10^{-3}$ and decays to $10^{-5}$ using a ReduceLROnPlateau scheduler (factor $0.5$, patience $200$). A linear warmup of $500$ epochs is applied at the start of training.

Early stopping is used with a patience of $500$ epochs and minimum improvement threshold $\Delta = 10^{-14}$.

Training uses full-batch optimization: all mesh nodes for each data point are loaded simultaneously, and no minibatching is used.

\subsection{Loss Function}

The total loss combines reconstruction and regularization

\[
\mathcal{L}_{\text{total}}
=
\mathcal{L}_{\text{rec}}
+
0.3\,\mathcal{L}_{\text{reg}}
\]

The reconstruction loss is the mean squared error between predicted and ground-truth displacement fields

\[
\mathcal{L}_{\text{rec}}
=
\frac{1}{m}
\sum_{j=1}^{m}
\|
\hat{\mathbf{F}}_j - \mathbf{F}_j
\|_2^2
\]
The regularization coefficient is set to $\lambda = 0.3$.

\section{Coordinate Normalization for Ring Geometry}
\label{apx: coordinate}

Given a physical point $(x,y,z)$, the goal is to map it into a normalized ``unit-space'' representation that preserves angle, normalizes height, and remaps the radial coordinate into a designated band for each ring.

\subsection{Convert to Cylindrical Coordinates}

Let
\[
r = \sqrt{x^2+y^2}, \qquad
\theta = \operatorname{atan2}(y,x), \qquad
z = z.
\]

Only the radial coordinate $r$ and height $z$ are remapped; the angular coordinate $\theta$ is preserved.

\subsection{Normalize the Radial Coordinate Within a Ring}

Each ring is defined by:
\begin{itemize}
    \item centerline radius $c$,
    \item thickness $t$.
\end{itemize}

Its physical radial extent is therefore
\[
r_{\text{inner}} = c - \frac{t}{2}, \qquad
r_{\text{outer}} = c + \frac{t}{2}.
\]

For a point lying in that ring, the physical radius $r$ is first normalized across the ring thickness,
\[
\lambda = \frac{r - r_{\text{inner}}}{t},
\]
so that $\lambda = 0$ at the inner edge and $\lambda = 1$ at the outer edge.

This normalized value is then mapped into a prescribed unit-space radial band $[r_{\min}, r_{\max}]$:
\[
r_u = r_{\min} + (r_{\max} - r_{\min})\,\lambda
\]
or equivalently,
\[
r_u = r_{\min} + (r_{\max} - r_{\min})\frac{r - (c - t/2)}{t}.
\]

The bands are defined as follows:
\begin{itemize}
    \item \textbf{Outer ring:} $[r_{\min}, r_{\max}] = [0.75, 1.00]$
    \item \textbf{Inner ring:} $[r_{\min}, r_{\max}] = [0.25, 0.50]$
\end{itemize}

The interval $[0.50, 0.75]$ is intentionally left unused, corresponding to the excluded spoke region.

Thus:
\begin{itemize}
    \item the inner edge of the outer ring maps to $r_u = 0.75$,
    \item the outer edge of the outer ring maps to $r_u = 1.00$,
    \item the inner ring maps analogously into $[0.25, 0.50]$.
\end{itemize}

\subsection{Normalize the Vertical Coordinate}

Let $H$ denote the model height. The vertical coordinate is mapped linearly from $[0,H]$ to $[0,1]$:
\[
z_u = \frac{z}{H}.
\]

\subsection{Convert Back to Cartesian Coordinates in Unit Space}

The normalized point is stored in Cartesian form while preserving the original angle $\theta$:
\[
\mathbf{p}_u =
\begin{pmatrix}
r_u \cos\theta \\
r_u \sin\theta \\
z_u
\end{pmatrix}.
\]

Since $\cos\theta = x/r$ and $\sin\theta = y/r$, this can also be written without explicitly computing $\theta$:
\[
\mathbf{p}_u =
\begin{pmatrix}
r_u \dfrac{x}{\sqrt{x^2+y^2}} \\
r_u \dfrac{y}{\sqrt{x^2+y^2}} \\
\dfrac{z}{H}
\end{pmatrix}.
\]

\subsection{Full Transformation}

For a physical point $(x,y,z)$ belonging to a ring with parameters $(c,t,r_{\min},r_{\max},H)$, the complete mapping is
\[
\mathbf{p}_u =
\begin{pmatrix}
\left[
r_{\min} + (r_{\max}-r_{\min})
\frac{\sqrt{x^2+y^2}-(c-t/2)}{t}
\right]
\frac{x}{\sqrt{x^2+y^2}}
\\[1.2em]
\left[
r_{\min} + (r_{\max}-r_{\min})
\frac{\sqrt{x^2+y^2}-(c-t/2)}{t}
\right]
\frac{y}{\sqrt{x^2+y^2}}
\\[1.2em]
\dfrac{z}{H}
\end{pmatrix}.
\]

The factors $x/r$ and $y/r$ are simply $\cos\theta$ and $\sin\theta$, so this form preserves the original angular position without requiring an explicit trigonometric conversion.



\end{document}